\pdfoutput=1

\documentclass[11pt]{article}

\usepackage[]{acl}

\usepackage{times}
\usepackage{latexsym}

\usepackage[T1]{fontenc}

\usepackage[utf8]{inputenc}

\usepackage{microtype}

\usepackage{multirow}

\usepackage{amsmath}
\usepackage{tabularx}
\usepackage{algorithmicx}
\usepackage{algpseudocode}
\usepackage{amssymb}
\usepackage{color}
\usepackage{booktabs}
\usepackage{multirow}
\usepackage{balance}
\usepackage{listings}
\usepackage{lstautogobble}
\usepackage{multicol}
\usepackage{bold-extra}
\usepackage[ruled,vlined,linesnumbered,noresetcount]{algorithm2e}
\usepackage{textcomp}
\usepackage[normalem]{ulem}
\usepackage{footnote}
\usepackage{graphicx}
\usepackage{subfigure}

\newcommand\figref[1]{Figure~\ref{fig:#1}}

\newcommand\tabref[1]{Table~\ref{tab:#1}}

\title{Characterizing the Efficiency vs. Accuracy Trade-off for\\
Long-Context NLP Models}

\author{Phyllis Ang\\
  \normalsize{Duke University}\\
  \normalsize{Durham, North Carolina, USA}\\
  \small{phyllis.ang@duke.edu}\\\And
  Bhuwan Dhingra\\
  \normalsize{Duke University}\\
  \normalsize{Durham, North Carolina, USA}\\
  \small{bdhingra@cs.duke.edu}\\\And
  Lisa Wu Wills\\
  \normalsize{Duke University}\\
  \normalsize{Durham, North Carolina, USA}\\
  \small{lisa@cs.duke.edu}\\
}

\begin{document}
\maketitle
\begin{abstract}
  With many real-world applications of Natural Language Processing (NLP) comprising of long texts, there has been a rise in NLP benchmarks that measure the accuracy of models that can handle longer input sequences. 
  However, these benchmarks do not consider the trade-offs between accuracy, speed, and power consumption as input sizes or model sizes are varied.
  In this work, we perform a systematic study of this accuracy vs. efficiency trade-off on two widely used long-sequence models -- Longformer-Encoder-Decoder (LED) and Big Bird -- during fine-tuning and inference on four datasets from the SCROLLS benchmark. 
  To study how this trade-off differs across hyperparameter settings, we compare the models across four sequence lengths ($1024$, $2048$, $3072$, $4096$) and two model sizes (base and large) under a fixed resource budget.
  We find that LED consistently achieves better accuracy at
  lower energy costs than Big Bird.
  For summarization, we find that increasing model size is more energy efficient than increasing sequence length for higher accuracy.
  However, this comes at the cost of a large drop in inference speed.
  For question answering, we find that smaller models are both more efficient and more
  accurate due to the larger training batch sizes possible under a fixed resource budget.
\end{abstract}

\section{Introduction}
    Over the past few years, advances in sequence modeling have led to impressive results on several NLP benchmarks \citep{wang2019superglue, wang_benchmarking_2020}. 
    A closer look at these results reveals that higher accuracies are typically achieved by increasingly larger and computationally intensive models, which
    have large carbon footprints that can have an adverse effect on the environment \citep{strubell-etal-2019-energy}.
    
    This has led to the Green AI initiative, which urges researchers to consider energy and computational efficiency when evaluating models in order to promote those which achieve high accuracies with smaller carbon footprints \citep{schwartz_green_2019}.  
    However, although it has been a few years since Green AI was introduced, efficiency metrics have still not been integrated into many recently proposed benchmarks such as the Long Range Arena (LRA) \citep{tay_long_2020} and SCROLLS \citep{shaham_scrolls_2022}.
    These benchmarks serve as a strong basis for comparison between Transformer models in terms of accuracy.
    However, improved accuracy is often obtained by either increasing
    the input sequence length or the model size, and the energy
    cost of these improvements is not clear.
    Moreover, previous characterizations of model efficiency in terms of speed
    (e.g., in LRA)
    only focus on \textit{inter}-model comparisons, keeping model
    sizes and input sequence lengths fixed.
    Here, we argue that the accuracy-vs-efficiency trade-off also has implications
    for \textit{intra}-model comparisons when selecting hyperparameters -- e.g.,
    increasing the sequence length might positively impact accuracy but may also negatively impact efficiency metrics.
    As a result, when faced with a fixed resource budget, it is not clear whether
    practitioners
    should opt for increasing the model size or increasing the input
    length for the most efficient use of resources.
    
    In this work, we perform a systematic study of
    the trade-off between efficiency and accuracy for two widely used long-context
    NLP models -- Big Bird \cite{zaheer_big_2021} and Longformer-Encoder-Decoder (LED) \cite{beltagy_longformer_2020} -- on four datasets from the SCROLLS
    benchmark.\footnote{
    Code available at \url{https://github.com/phyllisayk/nlp-efficiency-tradeoff}.}
    We characterize efficiency using several metrics,
    including the total energy consumption during training,
    training speed, inference speed, and power efficiency.
    We compare the models across several different input lengths and
    two different model sizes (base and large).
    Overall, for summarization, we find that, perhaps surprisingly, increasing model 
    size is a more energy efficient way of increasing accuracy as compared to increasing
    sequence length.
    However, if inference speed is the main efficiency metric of interest, then smaller
    models should be preferred.
    For question answering, on the other hand, we find that using smaller models
    is more efficient in terms of all metrics \textit{and} more accurate due to the larger training batch sizes allowed
    under a fixed resource budget.

\section{Background}

      \subsection{NLP Benchmarks}
          Benchmarks such as SuperGLUE \citep{wang2019superglue} and SQuAD \citep{rajpurkar-etal-2018-know} have served as the gold standard in the development of NLP models. 
          However, these benchmarks only capture model performance on short text sequences while many NLP tasks of interest, such as question answering and summarization, involve long contexts. 
          Recently, several efficient Transformer models have been introduced which require sub-quadratic memory and time complexity with respect to the input length \cite{tay_efficient_2020}.
          Consequently, new standardized benchmarks have been introduced specifically
          focusing on the long sequence modeling capabilities of these models,
          including the Long Range Arena (LRA) \citep{tay_long_2020} and SCROLLS \citep{shaham_scrolls_2022}.
          
          Although LRA evaluates long-sequence models, it only contains two language datasets which artificially elongate the input sequences through byte tokenization.
          The SCROLLS benchmark, on the other hand,
          focuses on language tasks which naturally require synthesizing information from long sequences, including summarization, question answering, and classification.
          SCROLLS does not compare models in terms of efficiency at all, and
          while LRA compares model speeds, it only does so across different model architectures, ignoring the impact of hyperparameter choices.
          For our analysis, we utilize three summarization tasks and one question answering task from SCROLLS.
    
      \subsection{Energy Considerations}
          As deep learning models grow more complex to meet increasing demands, the computation required to run these models generates an increasingly larger energy cost \citep{strubell-etal-2019-energy}. 
          This has led to the Green AI initiative \citep{schwartz_green_2019} which demands higher energy efficiency while maintaining state-of-the-art accuracies. 
          A benchmark of the performance and energy efficiency of AI accelerators has been performed during training, but it only examined 2-layer LSTMs and vanilla Transformers \citep{wang_benchmarking_2020}. 
          HULK \cite{zhou_hulk_2021} is an NLP benchmark that evaluates the energy efficiency of several Transformer models (e.g., BERT \cite{devlin-etal-2019-bert} and RoBERTa \cite{liu2019roberta}) during pre-training, fine-tuning, and inference, but it does not consider long-range models.
          Additionally, neither of the benchmarks consider the effects of different sequence lengths on both energy efficiency and accuracy. 
          However, we confirm the observation from HULK that larger model sizes do not always
          imply lower efficiency.

\section{Methodology}
    Our main contribution is an analysis of
    how different sequence lengths affect the
    trade-off between accuracy, power, and speed in long-context Transformer models
    during fine-tuning and inference.
    Since our focus is on long-context NLP tasks,
    we investigated the following four input sequence lengths: $1024$, $2048$, $3072$, and $4096$. 
    
    \subsection{Datasets}
    \label{sec:datasets}
    We conduct our analyses on four datasets from the SCROLLS benchmark: GovReport \citep{huang2021govreport}, SummScreenFD \citep{chen_summscreen_2021}, QMSum \citep{zhong2021qmsum}, and Qasper \citep{dasigi2021qasper}.
    These datasets span two different tasks -- summarization and question answering -- which frequently
    involve long inputs.
    We provide a summary of these datasets in Table \ref{tab:dataset_stats} with more details provided in Appendix \ref{sec:appendix_scrolls_dataset}.
    We cast these datasets in a unified sequence-to-sequence format using the same procedure as done
    in SCROLLS.
    
    \begin{table}
        \centering
        \begin{tabular}{lcc}
        \hline
        \textbf{Dataset} & \textbf{Task} & \textbf{Avg Input Length}\\
        \hline
        GovReport   & Summ  & 7,897  \\
        SumScreenFD & Summ  & 5,639\\
        QMSum       & Summ  & 10,396 \\ 
        Qasper      & QA    & 3,671\\ \hline
        \end{tabular}
        \caption{An overview of the datasets from SCROLLS that were used in this paper. This is an abbreviated version of the table shown in the original SCROLLS paper \citep{shaham_scrolls_2022}. \textit{Summ} indicates summarization and \textit{QA} indicates Question Answering.
        See Appendix \ref{sec:appendix_scrolls_dataset} for more information. }
        \label{tab:dataset_stats}
    \end{table}
    
    \subsection{Models}
    \label{sec:models}
    Following standard practice, we start with pretrained models and restrict our analysis to the fine-tuning and inference stages.
    Since our tasks are cast in a sequence-to-sequence format, we
    pick two widely used encoder-decoder models for long-context NLP -- the Longformer-Encoder-Decoder (LED) and Big Bird.
    To mimic a typical use-case, we obtained these two pre-trained models from the HuggingFace library\footnote{\url{https://huggingface.co/}} -- hence our analysis can be easily
    extended to any HuggingFace model.
    
    \paragraph{Longformer-Encoder-Decoder (LED).} 
    We analyzed both the base and large version of the LED model released with the original paper \citep{beltagy_longformer_2020}.
    This version of the LED model utilized the \verb;Longformer-chunks; implementation that achieves high compute efficiency at the cost of higher memory by chunking the key and query matrices such that only a single matrix multiplication operation from PyTorch is needed.
    The two versions of the model are stored on HuggingFace as {\small \verb;allenai/led-base-16384;} and  {\small \verb;allenai/led-large-16384;}. 
    
    \paragraph{Big Bird.} 
    Following the encoder-decoder setup in the original Big Bird paper \citep{zaheer_big_2021}, we utilized the version of Big Bird-large that has been pretrained on the PubMed dataset starting from Pegasus-large. 
    This model is stored on HuggingFace as {\small \verb;google/bigbird-pegasus-large-pubmed;}.
    We only performed experiments on the large version of this model as the base version is not released on HuggingFace.
    
    \subsection{Hardware Resources Provisioned}

    \begin{table*}
        \centering
        \begin{tabular}{lccccccc}
        \hline
        \multirow{2}{*}{\textbf{Dataset}} & \multirow{2}{*}{\textbf{Seq Len}} & \multicolumn{2}{c}{\textbf{1 GPU}} & \multicolumn{2}{c}{\textbf{1 GPU - Accum}} & \multicolumn{2}{c}{\textbf{8 GPUs - Accum}} \\
        & & \textbf{Batch Size} & \textbf{Acc} & \textbf{Batch Size} & \textbf{Acc} & \textbf{Batch Size} & \textbf{Acc} \\
        \hline
        \multirow{4}{*}{Qasper} & 1024 & 24 & 17.68 & 96 & 21.39 & 704 & 25.30 \\
        & 2048 & 12 & 22.74 & 48 & 27.87 & 352 & 29.97 \\
        & 3072 & 8 & 29.57 & 32 & 33.75 & 224 & 33.94  \\
        & 4096 & 6 & 32.88 & 24 & 34.20 & 160 & 36.36  \\
        \hline
        \multirow{4}{*}{GovReport} & 1024 & 24 & 49.53 & 96 & 49.53 & 704 & 48.78 \\
        & 2048 & 12 & 51.15 & 48 & 51.28 & 352 & 50.18 \\
        & 3072 & 8 & 51.67 & 32 & 52.09 & 224 & 50.60  \\
        & 4096 & 6 & 51.71 & 24 & 52.27 & 160 & 50.95  \\
        \hline
        \end{tabular}
        \caption{
        Accuracy of the LED-base model with varying batch sizes across different hardware configurations.
        \textit{Accum} indicates that a gradient accumulation step size of four was used to obtain the larger batch sizes.
        On the Qasper question answering task, where \textit{Acc} represents the F1 score of the predicted answers,
        increasing the batch sizes significantly improves the accuracy for all sequence lengths.
        On the GovReport summarization task, where \textit{Acc} represents the Rouge score,
        increasing the batch sizes has a negligible effect.}
        \label{tab:qasper_hardware}
    \end{table*}

    Our initial experiments with the LED-base model suggest that large
batch sizes are imperative for obtaining high accuracies on the
question answering task but less so for the summarization tasks
(see~\tabref{qasper_hardware}).  Quadrupling the batch sizes on the
Qasper question answering dataset -- through the use of gradient
accumulation step size of four -- resulted in a two to four point increase in the F1
scores across the input sequence lengths. Take the input sequence
length of $1024$ as an example (i.e., first row
of~\tabref{qasper_hardware}), we were able to fit a batch size of $24$
on one GPU (labeled \emph{1 GPU}) without suffering an out-of-memory
error when performing fine-tuning, obtaining a modest F1 score of
$17.68$. When we quadrupled the batch size to $96$ by using gradient
accumulation with step size of four (labeled \emph{1 GPU - Accum}), the model accuracy went
up to an F1 score of $21.39$.  When the batch sizes were further
increased through the use of more GPUs (labeled \emph{8 GPUs -
Accum}), the increase in F1 scores becomes more prominent at four to
seven points. The same trends hold for all sequence lengths on the
Qasper dataset.  On the other hand, quadrupling the batch sizes for
the GovReport summarization dataset resulted in negligible increases
in Rouge scores while the further increase via multiple GPUs actually
resulted in (slightly) lower Rouge scores.
    
    These initial experiments informed our decision to use a fixed
resource budget of $1$ Nvidia RTX A6000 GPU for both fine-tuning and
inference of all models on the summarization tasks, since increasing
the number of GPUs does not have a positive effect on the model
accuracy. On the other hand, for the question answering task, we used
a much larger fixed resource budget of $8$ Nvidia RTX A6000 GPUs (on
the same server) for both fine-tuning and inference to allow for larger batch sizes that can obtain much better model accuracy.
    
    \subsection{Fine-tuning}
    \label{sec:method_fine-tuning}
    
    
    All pre-trained models mentioned in Section \ref{sec:models} are fined-tuned without mixed precision or gradient checkpointing on all datasets until convergence.
    A model has converged when the accuracy metric of interest for that specific task stays the same or has worsened for $3$ validation calls. 
    In our case, since we perform validation every $500$ steps for summarization tasks and every $10$ steps for the question answering task, a model has converged when the metric has stayed the same or worsened for $1500$ steps for summarization tasks and $30$ steps for the question answering task.
    
    In terms of hyperparameters, we used the same hyperparameters that the SCROLLS benchmark utilized for the LED-base model except for the batch sizes.
    To control for the effects of memory on our metrics, for each sequence length and model, we selected the largest batch size that can fit on the $48$GB A6000 GPU.
    For the question answering task, the batch sizes were selected so that the minibatches on each of the $8$ GPUs were maximized.
    To further increase the effective size of each of minibatches in the question answering task, we set gradient accumulation steps to four.
    More information about the hyperparameters is outlined in Appendix \ref{sec:appendix_scrolls_hyperparam}.


    \subsection{Inference}
    Since we do not have access to the labels in the test sets of SCROLLS, inference is run on the validation set using the fine-tuned models. 
    All of our inferences were performed with a batch size of 16. 
    
    \subsection{Evaluation Criteria}
    \paragraph{Accuracy.} Our evaluation metrics for accuracy of the models on each dataset follow those mentioned in the SCROLLS paper. 
    GovReport, SummScreenFD, and QMSum are evaluated using Rouge, as is standard for summarization; Qasper is evaluated using a token-level F1 score after normalizing both the predicted
    and ground-truth answer strings.\footnote{Normalization is done in the same manner as Squad \cite{rajpurkar-etal-2018-know}).}
    For Rouge, following SCROLLS, we calculated the geometric mean of three different types of rouge to provide a single value: Rouge-1 (unigram overlap), Rouge-2 (bigram overlap), and Rouge-L (longest sequence overlap).
    
    \paragraph{Efficiency.} For efficiency metrics, we explored the training power efficiency (number of samples trained per second per Watt), total training energy required (average power $\times$ training time), training speed (number of samples trained per second), and inference speed (number of samples inferenced per second). 
    The training and inference speeds are provided by the HuggingFace library while the total energy consumed and the power efficiency of the GPU(s) were collected with the help of the Weights and Biases (wandb) tool.\footnote{\url{https://wandb.ai/site}}
    
    We chose power efficiency as one of our metrics because it is one of the most important industry standard metrics used for machine learning platforms (TPU uses performance per Watt, MLPerf~\citep{reddi_mlperf_2020, mattson_mlperf_2020} measures the number of samples inferenced per second per Watt) as it is a key component of TCO (Total Cost of Ownership). Cloud providers routinely spend 40-50\% of the cost towards electricity as well as powering and cooling the servers, and this cost is increasing.
    Hence, maximizing the utility of this spent power by increasing the number of samples processed
    per watt is crucial for reducing the carbon footprint of NLP research.
    

\begin{figure*}[!t]
  \centering
  \includegraphics[width=0.9\textwidth]{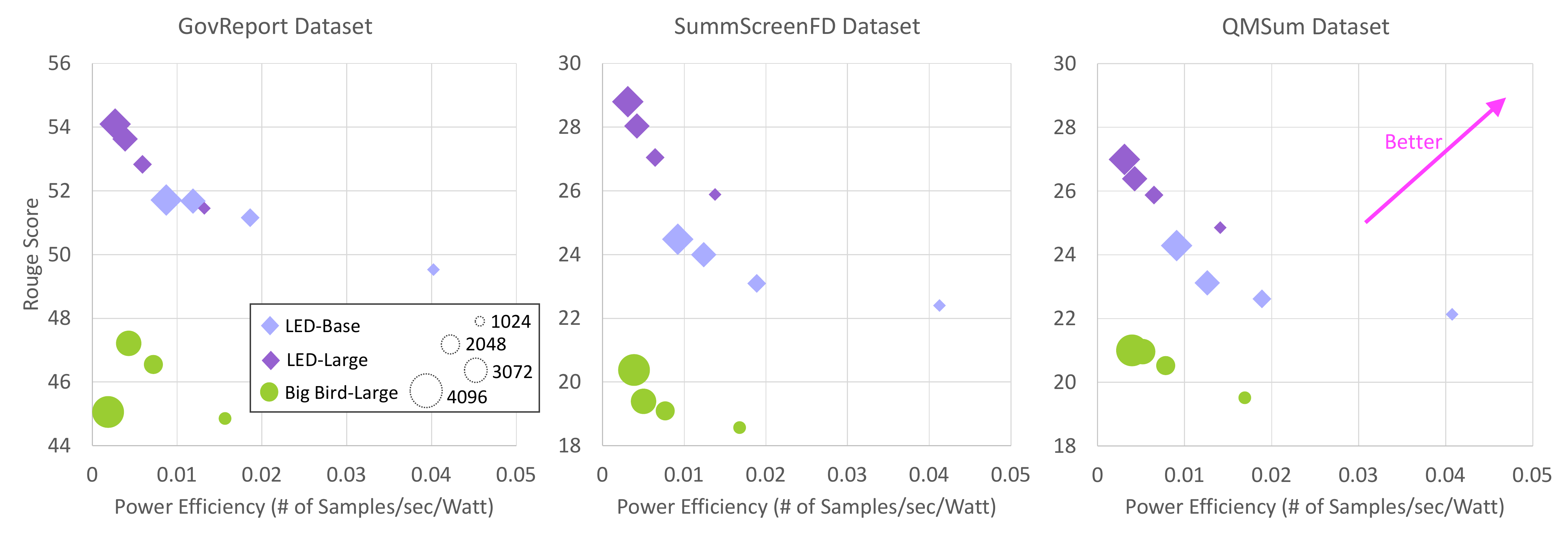}
  \caption{Power efficiency measured in number of samples per second per watt vs. model accuracy in Rouge score for the three summarization datasets -- GovReport (Left), SummScreenFD (Middle), QMSum (Right) -- while varying input sequence lengths.}
  \label{fig:power-eff-summ}
\end{figure*}


\begin{figure*}[!t]
  \centering
  \includegraphics[width=0.9\textwidth]{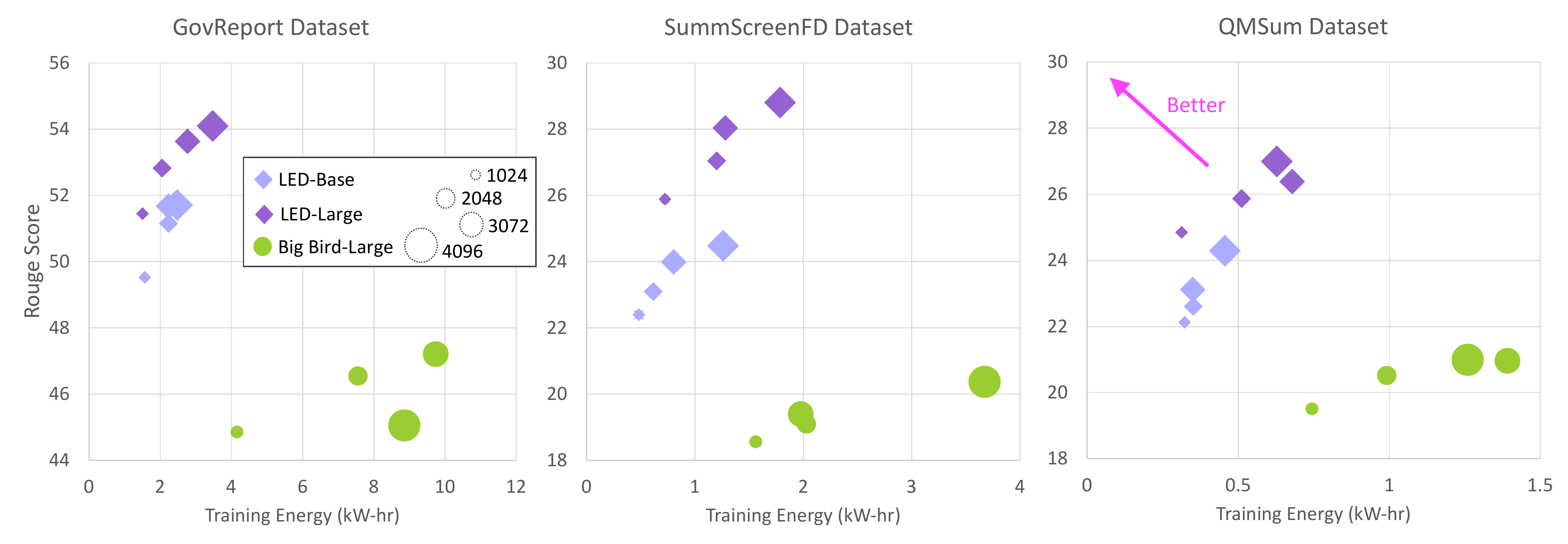}
  \caption{Total training energy consumption measured in kiloWatt-hour vs. model accuracy in Rouge score for the three summarization datasets -- GovReport (Left), SummScreenFD (Middle), QMSum (Right) -- while varying input sequence lengths.}
  \label{fig:energy-summ}
\end{figure*}

\section{Results}
\subsection{Summarization Datasets}
\figref{power-eff-summ} depicts the power efficiency of each summarization dataset vs. its corresponding training accuracy for input lengths ranging from $1024$ to $4096$ tokens. 
We make the following observations: First, power efficiency has a strong inverse correlation with the size of the input sequence lengths, with small variations across datasets.
Second, the Big Bird-large model has similar power efficiency to LED-large model across the input sequence lengths, but Big Bird's Rouge scores are much lower, making one of the LED models a better choice to select when training summarization tasks. 

\figref{energy-summ} shows the total energy consumed during training on each of the three summarization
datasets.
Interestingly, we observe that on GovReport and QMSum, LED-large with sequence length $1024$
is more efficient \textit{and} has higher accuracy than each of the LED-base models with larger
sequence lengths.
Increasing the sequence length for LED-large further increases this accuracy while
still often being more efficient than LED-base models with greater sequence lengths.
This suggests that, for summarization, using larger models with short sequence lengths
is a more energy friendly way to get higher accuracies (as compared to small models with larger
sequence lengths).
We find Big Bird to both consume more energy and achieve lower Rouge scores.


\begin{figure*}[!t]
  \centering
  \includegraphics[width=0.9\textwidth]{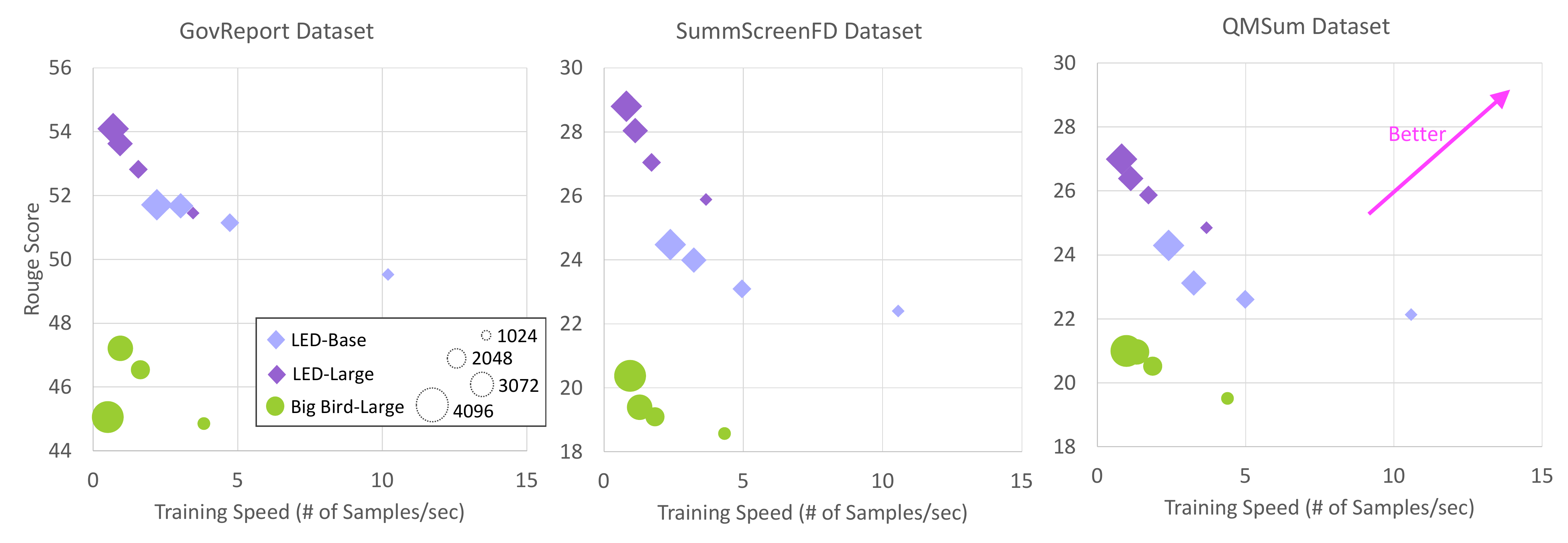}
  \caption{Model training speed measured in number of samples per second vs. model accuracy in Rouge score for the three summarization datasets -- GovReport (Left), SummScreenFD (Middle), QMSum (Right) -- while varying input sequence lengths.}
  \label{fig:training-speed-summ}
\end{figure*}


\begin{figure*}[!t]
  \centering
  \includegraphics[width=0.9\textwidth]{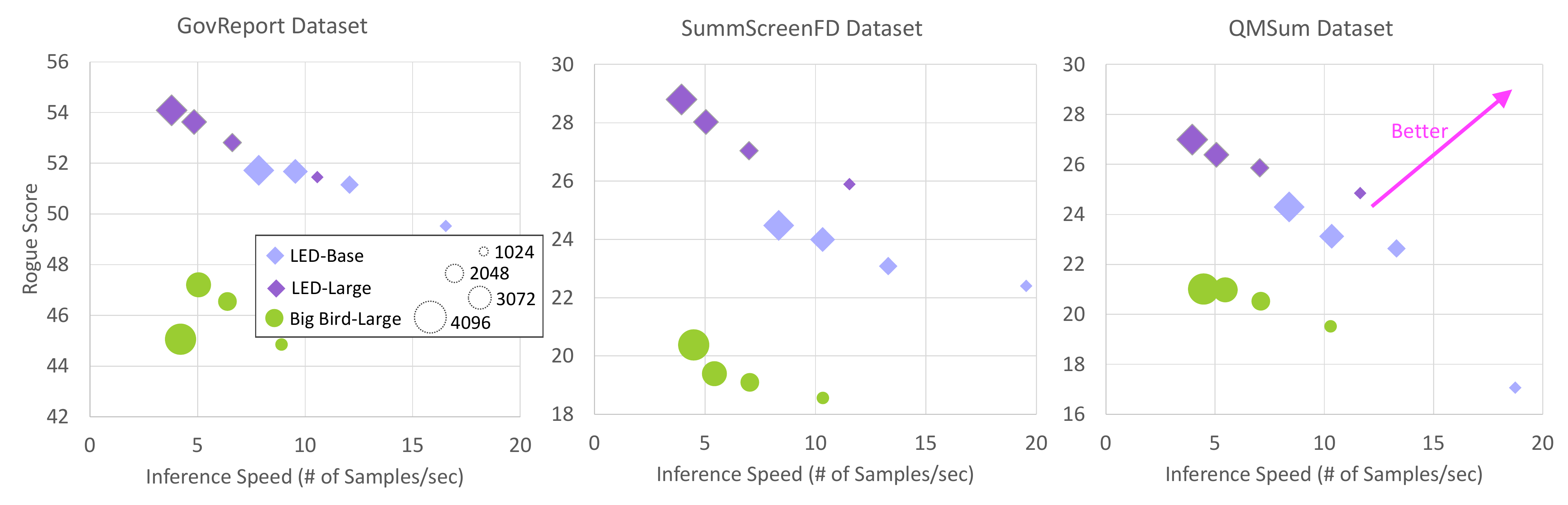}
  \caption{Model inference speed measured in number of samples per second vs. model accuracy in Rouge score for the three summarization datasets -- GovReport (Left), SummScreenFD (Middle), QMSum (Right) -- while varying input sequence lengths.}
  \label{fig:inference-speed-summ}
\end{figure*}

The training speed (\figref{training-speed-summ}) and the inference speed (\figref{inference-speed-summ}) of the summarization datasets show similar trends. As the input sequence lengths increase, the training and inference speeds decrease due to the sub-quadratic runtime complexity (with respect to the input sequence lengths) exhibited in the attention mechanisms employed in these efficient Transformer models. 
Unlike training energy, inference speed increases when the model size is smaller at the cost of lower accuracy.
However, sometimes (such as the datapoints exhibited in the GovReport dataset) a similar accuracy can be obtained by LED-base model with a larger input length ($2048$) as opposed to LED-large with a smaller input length ($1024$).

\subsection{Qasper Dataset and Scaling Up Resources}


\begin{figure*}[!t]
  \centering
  \includegraphics[width=1\textwidth]{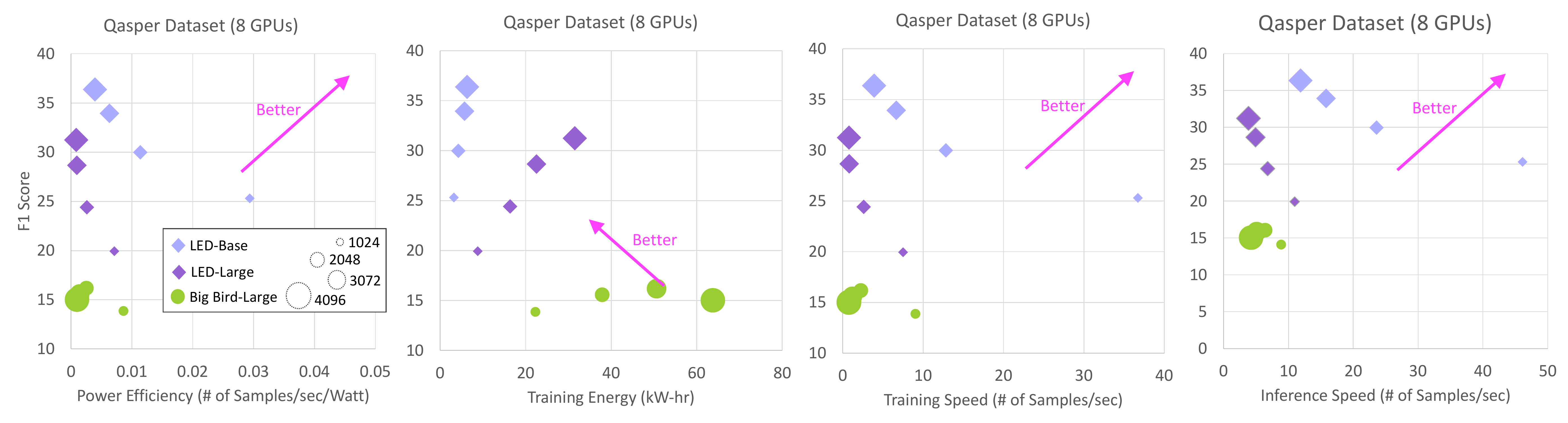}
  \caption{Power efficiency measured in number of samples per second (Left), training energy estimated in kiloWatt-hour (Center Left), training speed (Center Right) and inference speed (Right) in number of samples per second vs. model accuracy in F1 score for the Qasper question answering dataset while varying input sequence lengths.}
  \label{fig:qasper-everything}
\end{figure*}

\figref{qasper-everything} shows all four efficiency metrics for the Qasper question answering task.
Once again, the LED models outperform Big Bird in the overall F1 score.
Interestingly, we observe that under fixed resources, LED-base also outperforms
LED-large on this dataset.\footnote{
We note that our LED-base model with input sequence length $4096$
achieves an F1 score of approximately $10$ points higher than what was reported
in the SCROLLS paper.
}
We suspect this is due to the larger batch sizes we can fit for LED-base as compared
to LED-large, which we found to be particularly important for this dataset.
Hence, we found it to be more efficient and more accurate to use the smaller model on this task.
Increasing sequence length brings large gains in accuracy
with a small increased cost in training energy but a large slow-down in terms
of speed.

\subsection{Energy Consumption Deep Dive}

\begin{figure*}[!t]
  \centering
  \includegraphics[width=1\textwidth]{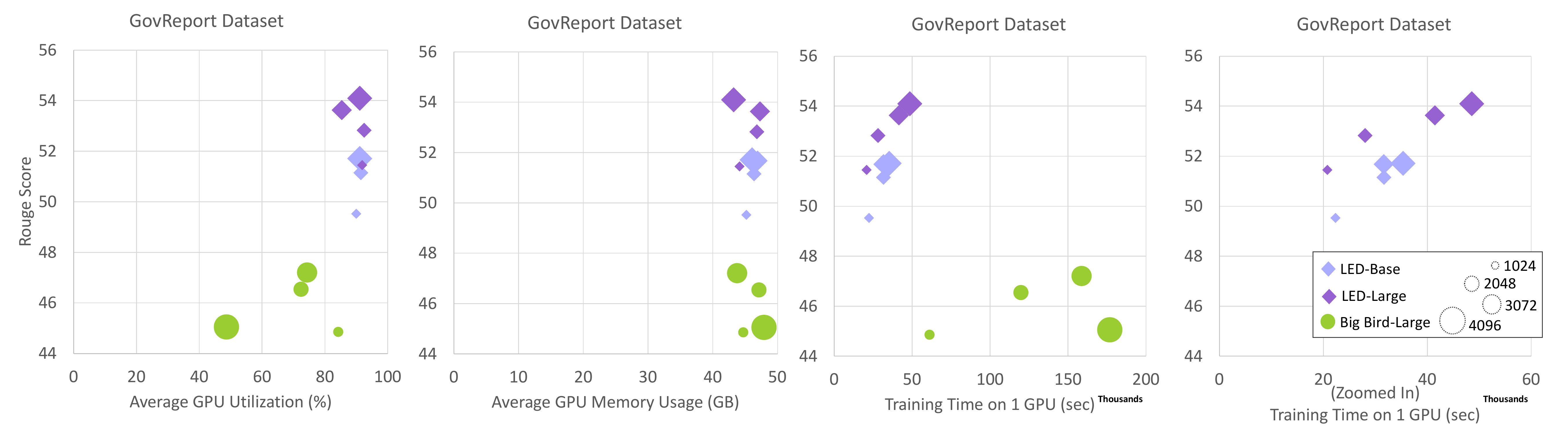}
  \caption{Average GPU utilization (Left), average GPU memory usage (Center Left), and total training time in seconds (Center Right and Right) vs. model accuracy for the GovReport summarization dataset while varying input sequence lengths.}  
  \label{fig:gov-power-deepdive}
\end{figure*}

To understand the energy consumption of the hardware platform, we present a deeper analysis on the GovReport dataset. We plot the GPU utilization (as an average over the entire training run), the GPU memory usage (as an average over the entire training run), and the training time (in seconds) in~\figref{gov-power-deepdive}. 
From the GPU utilization plot, we observe that the single GPU is pretty well utilized for the LED models while Big Bird seems to not saturate the GPU especially when the input sequence length is $4096$.  
This would suggest that Big Bird would incur a smaller energy cost because not all GPU resources are online.
However, Big Bird took about $48$ hours to train for a sequence length of $4096$ while LED-large took $14$ hours to train at the same sequence length. The almost four times in training time contributed to Big Bird's high energy consumption in~\figref{energy-summ}, making it the least carbon-friendly model to train for GovReport.
In general, the training time on the GPU (depicted in~\figref{gov-power-deepdive}-right) exhibits a similar trend as the total energy consumed.
The average GPU utilization is therefore not an indicative metric in predicting the energy consumption of model training in this case, but the training time is, as energy is calculated using power consumed over time (or the area under the curve when plotting power over time).

\section{Conclusion}

We have presented a systematic study of the accuracy vs. efficiency trade-offs
involved in four long-context NLP tasks across two model architectures.
In addition to comparing model architectures as commonly done in NLP benchmarks,
our focus was on comparing models of two different sizes and four different sequence lengths.
We highlight several key findings which we hope practitioners can utilize
to select hyperparameters under a resource constrained setting.
One such key finding is that using a larger model instead of larger input sequence lengths is a more energy friendly way to achieve higher accuracies on summarization tasks if inference speed is not a concern. 
On the other hand, utilizing a longer input sequence length with a smaller model for question answering task results in higher accuracies with higher efficiency.

\bibliography{2022-nlp_power-longseq}

\begin{thebibliography}{19}
\expandafter\ifx\csname natexlab\endcsname\relax\def\natexlab#1{#1}\fi

\bibitem[{Beltagy et~al.(2020)Beltagy, Peters, and
  Cohan}]{beltagy_longformer_2020}
Iz~Beltagy, Matthew~E. Peters, and Arman Cohan. 2020.
\newblock \href {http://arxiv.org/abs/2004.05150} {Longformer: {The}
  {Long}-{Document} {Transformer}}.
\newblock \emph{arXiv:2004.05150 [cs]}.
\newblock ArXiv: 2004.05150.

\bibitem[{Chen et~al.(2021)Chen, Chu, Wiseman, and
  Gimpel}]{chen_summscreen_2021}
Mingda Chen, Zewei Chu, Sam Wiseman, and Kevin Gimpel. 2021.
\newblock \href {http://arxiv.org/abs/2104.07091} {{SummScreen}: {A} {Dataset}
  for {Abstractive} {Screenplay} {Summarization}}.
\newblock \emph{arXiv:2104.07091 [cs]}.
\newblock ArXiv: 2104.07091.

\bibitem[{Dasigi et~al.(2021)Dasigi, Lo, Beltagy, Cohan, Smith, and
  Gardner}]{dasigi2021qasper}
Pradeep Dasigi, Kyle Lo, Iz~Beltagy, Arman Cohan, Noah~A. Smith, and Matt
  Gardner. 2021.
\newblock \href {https://doi.org/10.18653/v1/2021.naacl-main.365} {A dataset of
  information-seeking questions and answers anchored in research papers}.
\newblock In \emph{Proceedings of the 2021 Conference of the North American
  Chapter of the Association for Computational Linguistics: Human Language
  Technologies}, pages 4599--4610, Online. Association for Computational
  Linguistics.

\bibitem[{Devlin et~al.(2019)Devlin, Chang, Lee, and
  Toutanova}]{devlin-etal-2019-bert}
Jacob Devlin, Ming-Wei Chang, Kenton Lee, and Kristina Toutanova. 2019.
\newblock \href {https://doi.org/10.18653/v1/N19-1423} {{BERT}: Pre-training of
  deep bidirectional transformers for language understanding}.
\newblock In \emph{Proceedings of the 2019 Conference of the North {A}merican
  Chapter of the Association for Computational Linguistics: Human Language
  Technologies, Volume 1 (Long and Short Papers)}, pages 4171--4186,
  Minneapolis, Minnesota. Association for Computational Linguistics.

\bibitem[{Huang et~al.(2021)Huang, Cao, Parulian, Ji, and
  Wang}]{huang2021govreport}
Luyang Huang, Shuyang Cao, Nikolaus Parulian, Heng Ji, and Lu~Wang. 2021.
\newblock \href {https://doi.org/10.18653/v1/2021.naacl-main.112} {Efficient
  attentions for long document summarization}.
\newblock In \emph{Proceedings of the 2021 Conference of the North American
  Chapter of the Association for Computational Linguistics: Human Language
  Technologies}, pages 1419--1436, Online. Association for Computational
  Linguistics.

\bibitem[{Liu et~al.(2019)Liu, Ott, Goyal, Du, Joshi, Chen, Levy, Lewis,
  Zettlemoyer, and Stoyanov}]{liu2019roberta}
Yinhan Liu, Myle Ott, Naman Goyal, Jingfei Du, Mandar Joshi, Danqi Chen, Omer
  Levy, Mike Lewis, Luke Zettlemoyer, and Veselin Stoyanov. 2019.
\newblock Roberta: A robustly optimized bert pretraining approach.
\newblock \emph{arXiv preprint arXiv:1907.11692}.

\bibitem[{Mattson et~al.(2020)Mattson, Cheng, Diamos, Coleman, Micikevicius,
  Patterson, Tang, Wei, Bailis, Bittorf, Brooks, Chen, Dutta, Gupta, Hazelwood,
  Hock, Huang, Kang, Kanter, Kumar, Liao, Narayanan, Oguntebi, Pekhimenko,
  Pentecost, Janapa~Reddi, Robie, St~John, Wu, Xu, Young, and
  Zaharia}]{mattson_mlperf_2020}
Peter Mattson, Christine Cheng, Gregory Diamos, Cody Coleman, Paulius
  Micikevicius, David Patterson, Hanlin Tang, Gu-Yeon Wei, Peter Bailis, Victor
  Bittorf, David Brooks, Dehao Chen, Debo Dutta, Udit Gupta, Kim Hazelwood,
  Andy Hock, Xinyuan Huang, Daniel Kang, David Kanter, Naveen Kumar, Jeffery
  Liao, Deepak Narayanan, Tayo Oguntebi, Gennady Pekhimenko, Lillian Pentecost,
  Vijay Janapa~Reddi, Taylor Robie, Tom St~John, Carole-Jean Wu, Lingjie Xu,
  Cliff Young, and Matei Zaharia. 2020.
\newblock \href
  {https://proceedings.mlsys.org/paper/2020/file/02522a2b2726fb0a03bb19f2d8d9524d-Paper.pdf}
  {Mlperf training benchmark}.
\newblock In \emph{Proceedings of Machine Learning and Systems}, volume~2,
  pages 336--349.

\bibitem[{Rajpurkar et~al.(2018)Rajpurkar, Jia, and
  Liang}]{rajpurkar-etal-2018-know}
Pranav Rajpurkar, Robin Jia, and Percy Liang. 2018.
\newblock \href {https://doi.org/10.18653/v1/P18-2124} {Know what you don{'}t
  know: Unanswerable questions for {SQ}u{AD}}.
\newblock In \emph{Proceedings of the 56th Annual Meeting of the Association
  for Computational Linguistics (Volume 2: Short Papers)}, pages 784--789,
  Melbourne, Australia. Association for Computational Linguistics.

\bibitem[{Reddi et~al.(2020)Reddi, Cheng, Kanter, Mattson, Schmuelling, Wu,
  Anderson, Breughe, Charlebois, Chou, Chukka, Coleman, Davis, Deng, Diamos,
  Duke, Fick, Gardner, Hubara, Idgunji, Jablin, Jiao, John, Kanwar, Lee, Liao,
  Lokhmotov, Massa, Meng, Micikevicius, Osborne, Pekhimenko, Rajan, Sequeira,
  Sirasao, Sun, Tang, Thomson, Wei, Wu, Xu, Yamada, Yu, Yuan, Zhong, Zhang, and
  Zhou}]{reddi_mlperf_2020}
Vijay~Janapa Reddi, Christine Cheng, David Kanter, Peter Mattson, Guenther
  Schmuelling, Carole-Jean Wu, Brian Anderson, Maximilien Breughe, Mark
  Charlebois, William Chou, Ramesh Chukka, Cody Coleman, Sam Davis, Pan Deng,
  Greg Diamos, Jared Duke, Dave Fick, J.~Scott Gardner, Itay Hubara, Sachin
  Idgunji, Thomas~B. Jablin, Jeff Jiao, Tom~St John, Pankaj Kanwar, David Lee,
  Jeffery Liao, Anton Lokhmotov, Francisco Massa, Peng Meng, Paulius
  Micikevicius, Colin Osborne, Gennady Pekhimenko, Arun Tejusve~Raghunath
  Rajan, Dilip Sequeira, Ashish Sirasao, Fei Sun, Hanlin Tang, Michael Thomson,
  Frank Wei, Ephrem Wu, Lingjie Xu, Koichi Yamada, Bing Yu, George Yuan, Aaron
  Zhong, Peizhao Zhang, and Yuchen Zhou. 2020.
\newblock \href {http://arxiv.org/abs/1911.02549} {{MLPerf} {Inference}
  {Benchmark}}.
\newblock \emph{arXiv:1911.02549 [cs, stat]}.
\newblock ArXiv: 1911.02549.

\bibitem[{Schwartz et~al.(2020)Schwartz, Dodge, Smith, and
  Etzioni}]{schwartz_green_2019}
Roy Schwartz, Jesse Dodge, Noah~A. Smith, and Oren Etzioni. 2020.
\newblock \href {https://doi.org/10.1145/3381831} {Green ai}.
\newblock \emph{Commun. ACM}, 63(12):54–63.

\bibitem[{Shaham et~al.(2022)Shaham, Segal, Ivgi, Efrat, Yoran, Haviv, Gupta,
  Xiong, Geva, Berant, and Levy}]{shaham_scrolls_2022}
Uri Shaham, Elad Segal, Maor Ivgi, Avia Efrat, Ori Yoran, Adi Haviv, Ankit
  Gupta, Wenhan Xiong, Mor Geva, Jonathan Berant, and Omer Levy. 2022.
\newblock \href {http://arxiv.org/abs/2201.03533} {{SCROLLS}: {Standardized}
  {CompaRison} {Over} {Long} {Language} {Sequences}}.
\newblock \emph{arXiv:2201.03533 [cs, stat]}.
\newblock ArXiv: 2201.03533.

\bibitem[{Strubell et~al.(2019)Strubell, Ganesh, and
  McCallum}]{strubell-etal-2019-energy}
Emma Strubell, Ananya Ganesh, and Andrew McCallum. 2019.
\newblock \href {https://doi.org/10.18653/v1/P19-1355} {Energy and policy
  considerations for deep learning in {NLP}}.
\newblock In \emph{Proceedings of the 57th Annual Meeting of the Association
  for Computational Linguistics}, pages 3645--3650, Florence, Italy.
  Association for Computational Linguistics.

\bibitem[{Tay et~al.(2020{\natexlab{a}})Tay, Dehghani, Abnar, Shen, Bahri,
  Pham, Rao, Yang, Ruder, and Metzler}]{tay_long_2020}
Yi~Tay, Mostafa Dehghani, Samira Abnar, Yikang Shen, Dara Bahri, Philip Pham,
  Jinfeng Rao, Liu Yang, Sebastian Ruder, and Donald Metzler.
  2020{\natexlab{a}}.
\newblock \href {http://arxiv.org/abs/2011.04006} {Long {Range} {Arena}: {A}
  {Benchmark} for {Efficient} {Transformers}}.
\newblock \emph{arXiv:2011.04006 [cs]}.
\newblock ArXiv: 2011.04006.

\bibitem[{Tay et~al.(2020{\natexlab{b}})Tay, Dehghani, Bahri, and
  Metzler}]{tay_efficient_2020}
Yi~Tay, Mostafa Dehghani, Dara Bahri, and Donald Metzler. 2020{\natexlab{b}}.
\newblock \href {http://arxiv.org/abs/2009.06732} {Efficient {Transformers}:
  {A} {Survey}}.
\newblock \emph{arXiv:2009.06732 [cs]}.
\newblock ArXiv: 2009.06732.

\bibitem[{Wang et~al.(2019)Wang, Pruksachatkun, Nangia, Singh, Michael, Hill,
  Levy, and Bowman}]{wang2019superglue}
Alex Wang, Yada Pruksachatkun, Nikita Nangia, Amanpreet Singh, Julian Michael,
  Felix Hill, Omer Levy, and Samuel Bowman. 2019.
\newblock Superglue: A stickier benchmark for general-purpose language
  understanding systems.
\newblock \emph{Advances in neural information processing systems}, 32.

\bibitem[{Wang et~al.(2020)Wang, Wang, Shi, He, Tang, Zhao, and
  Chu}]{wang_benchmarking_2020}
Y.~Wang, Q.~Wang, S.~Shi, X.~He, Z.~Tang, K.~Zhao, and X.~Chu. 2020.
\newblock \href {https://doi.org/10.1109/CCGrid49817.2020.00-15} {Benchmarking
  the {Performance} and {Energy} {Efficiency} of {AI} {Accelerators} for {AI}
  {Training}}.
\newblock In \emph{2020 20th {IEEE}/{ACM} {International} {Symposium} on
  {Cluster}, {Cloud} and {Internet} {Computing} ({CCGRID})}, pages 744--751.

\bibitem[{Zaheer et~al.(2020)Zaheer, Guruganesh, Dubey, Ainslie, Alberti,
  Ontanon, Pham, Ravula, Wang, Yang, and Ahmed}]{zaheer_big_2021}
Manzil Zaheer, Guru Guruganesh, Kumar~Avinava Dubey, Joshua Ainslie, Chris
  Alberti, Santiago Ontanon, Philip Pham, Anirudh Ravula, Qifan Wang, Li~Yang,
  and Amr Ahmed. 2020.
\newblock \href
  {https://proceedings.neurips.cc/paper/2020/file/c8512d142a2d849725f31a9a7a361ab9-Paper.pdf}
  {Big bird: Transformers for longer sequences}.
\newblock In \emph{Advances in Neural Information Processing Systems},
  volume~33, pages 17283--17297. Curran Associates, Inc.

\bibitem[{Zhong et~al.(2021)Zhong, Yin, Yu, Zaidi, Mutuma, Jha, Awadallah,
  Celikyilmaz, Liu, Qiu, and Radev}]{zhong2021qmsum}
Ming Zhong, Da~Yin, Tao Yu, Ahmad Zaidi, Mutethia Mutuma, Rahul Jha,
  Ahmed~Hassan Awadallah, Asli Celikyilmaz, Yang Liu, Xipeng Qiu, and Dragomir
  Radev. 2021.
\newblock \href {https://doi.org/10.18653/v1/2021.naacl-main.472} {{QMS}um: A
  new benchmark for query-based multi-domain meeting summarization}.
\newblock In \emph{Proceedings of the 2021 Conference of the North American
  Chapter of the Association for Computational Linguistics: Human Language
  Technologies}, pages 5905--5921, Online. Association for Computational
  Linguistics.

\bibitem[{Zhou et~al.(2021)Zhou, Chen, Jin, and Wang}]{zhou_hulk_2021}
Xiyou Zhou, Zhiyu Chen, Xiaoyong Jin, and William~Yang Wang. 2021.
\newblock \href {https://doi.org/10.18653/v1/2021.eacl-demos.39} {{HULK}: {An}
  {Energy} {Efficiency} {Benchmark} {Platform} for {Responsible} {Natural}
  {Language} {Processing}}.
\newblock In \emph{Proceedings of the 16th {Conference} of the {European}
  {Chapter} of the {Association} for {Computational} {Linguistics}: {System}
  {Demonstrations}}, pages 329--336, Online. Association for Computational
  Linguistics.

\end{thebibliography}

\appendix

\section{SCROLLS Dataset}
\label{sec:appendix_scrolls_dataset}
    Table~\ref{tab:scrolls_overview} gives an overview of the datasets used in this paper,
    and we provide a brief description of each dataset below.
    
    \begin{table*}
    \centering
    \begin{tabular}{lllllll}
    \hline
        \multirow{2}{*}{\textbf{Dataset}} & \multirow{2}{*}{\textbf{Task}} & \multirow{2}{*}{\textbf{Domain}} & \multirow{2}{*}{\textbf{Metric}} & \multicolumn{2}{c}{\textbf{Avg \#Words}} &   \multirow{2}{*}{\textbf{\#Examples}}\\
        &&&& \textbf{Input} & \textbf{Output}\\
    \hline
        GovReport & Summ & Government & ROUGE & 7,897 & 492.7 & 19,402 \\
        SummScreenFD & Summ & TV & ROUGE & 5,639 & 100.0 & 4,348 \\
        QMSum & QB-Summ & Meetings & ROUGE & 10,396 & 69.7 & 1,810 \\
        Qasper & QA & Science & F1 & 3,671 & 11.5 & 5,692 \\
    \hline
    \end{tabular}
    \caption{\label{tab:scrolls_overview}
    An overview of the datasets the SCROLLS dataset with their statistics that was recreated from the original SCROLLS paper \citep{shaham_scrolls_2022}. \textit{Summ} indicates summarization, \textit{QB-Summ} means query-based summarization and \textit{QA} means question answering.
    The number of examples for each dataset includes all the examples from train, validation, and test sets.
    }
    \end{table*}
    
    \paragraph{GovReport.} \citep{huang2021govreport} A summarization dataset comprised of reports published by the U.S. Government Accountability Office (GAO) and Congressional Research Service (CRS).
    
    \paragraph{SummScreenFD.} \citep{chen_summscreen_2021} A summarization dataset where the goal is to generate a summary of an episode of a TV show when given a transcript of the episode. 
    
    \paragraph{QMSum.} \citep{zhong2021qmsum} A query-based summarization dataset composed of meeting notes from various sources such as academic group meetings, industrial product meetings, and public policy meetings. Models have to be able summarize specific sections of meetings when given a query.
    
    \paragraph{Qasper.} \citep{dasigi2021qasper} A question answering dataset over NLP papers from Semantic Scholar Open Research Corpus (S2ORC). 
    Given the title and abstract of a paper, models have to be able to generate the answer to a question about the paper.
    
    
    


\section{SCROLLS Model Hyperparameters}
\label{sec:appendix_scrolls_hyperparam}
    All the experiments conducted in this project were built upon the pre-trained models from the HuggingFace library. 
    Many of the hyperparameters used here are the same as those used for the LED-base model in SCROLLS.
    Unless specified in Table \ref{tab:hyperparameters}, hyperparameters take on default values from the HuggingFace Trainer library.\footnote{
    \url{https://huggingface.co/docs/transformers/main_classes/trainer}}
    
    \begin{table}
        \centering
        \begin{tabular}{lc}
        \hline
        \textbf{Hyperparameter} & \textbf{Value} \\
        \hline
        Validation Accumulation Steps       & 10 \\
        Learning Rate (all other dataset)   & 2e-5 \\
        Learning Rate Scheduler             & Linear \\
        Learning Rate Warm-up Ratio         & 0.1 \\
        Adam Optimizer Epsilon              & 1e-6 \\ 
        Adam Optimizer Beta1                & 0.9\\ 
        Adam Optimizer Beta2                & 0.98\\
        Dataloader Workers                  & 1 \\
        Maximum Epoch                       & 50 \\
        Early Stopping                      & 3 \\
        \hline
        \end{tabular}
        \caption{Hyperparameters used during fine-tuning of the pre-trained models. For any hyperparameters that are not listed in this table, we used the default values provided from the HuggingFace Trainer Library \protect\footnotemark.}
        \label{tab:hyperparameters}
    \end{table}
    \footnotetext{See previous note.}
    
    As mentioned in Section \ref{sec:method_fine-tuning}, we selected the largest batch sizes that can fit on the NVIDIA RTX A6000 GPU(s) during fine-tuning for each model and dataset in order to control for the effects of memory on our metrics. 
    Table \ref{tab:batch_sizes} shows the batch sizes used for fine-tuning each model on the different datasets at different input sequence lengths.
    
    \begin{table}
        \centering
        \begin{tabular}{lccc}
        \hline
        \textbf{Task} & \textbf{Model} & \textbf{Seq Len} & \textbf{Batch} \\
        \hline
        \multirow{12}{*}{Summ}  & \multirow{4}{*}{LED-base} & 1024 & 24 \\
        & & 2048 & 12\\
        & & 3072 & 8 \\
        & & 4096 & 6 \\
        \cline{2-4}
        & \multirow{4}{*}{LED-large} & 1024 & 8 \\
        & & 2048 & 4\\
        & & 3072 & 3 \\
        & & 4096 & 2 \\
        \cline{2-4}
        & \multirow{4}{*}{Big Bird-large} & 1024 & 7 \\
        & & 2048 & 4\\
        & & 3072 & 2 \\
        & & 4096 & 2 \\
        \hline
        \multirow{12}{*}{QA}  & \multirow{4}{*}{LED-base} & 1024 & 704 \\
        & & 2048 & 352 \\
        & & 3072 & 224 \\
        & & 4096 & 160 \\
        \cline{2-4}
        & \multirow{4}{*}{LED-large} & 1024 & 256 \\
        & & 2048 & 128 \\
        & & 3072 & 64 \\
        & & 4096 & 64 \\
        \cline{2-4}
        & \multirow{4}{*}{Big Bird-large} & 1024 & 224 \\
        & & 2048 & 96\\
        & & 3072 & 64 \\
        & & 4096 & 32 \\
        \hline
        \end{tabular}
        \caption{Batch sizes used for fine-tuning the different models for each of the tasks at each input sequence length. \textit{Summ} indicates summarization, and \textit{QA} means question answering. The batch sizes listed for the QA task is the total batch size across the $8$ GPUs with gradient accumulation step set to four.}
        \label{tab:batch_sizes}
    \end{table}

\end{document}